\documentclass{article}
\usepackage{ijcai19}

\usepackage{times}
\usepackage{soul}
\usepackage{url}
\usepackage[hidelinks]{hyperref}
\usepackage[utf8]{inputenc}
\usepackage[small]{caption}
\usepackage{graphicx}
\usepackage{amsmath}
\usepackage{booktabs}
\usepackage{latexsym} 
\usepackage{multirow}
\usepackage{amssymb}
\usepackage{color}
\usepackage[linesnumbered,ruled]{algorithm2e}
\usepackage{subfigure}

\usepackage{cleveref}[2012/02/15]
\crefformat{footnote}{#2\footnotemark[#1]#3}

\title{Learning Word Embeddings with Domain Awareness}

\author{
Guoyin Wang$^{1\dag}$\footnote{Work was done during the author's internship in Tencent AI Lab} \and
Yan Song$^2$\footnote{Contact authors.} \and
Yue Zhang$^3$ \And
Dong Yu$^{2}$  
\\
\affiliations
$^1$Duke University\\
$^2$Tencent AI Lab\\
$^3$Westlake University
\emails
guoyin.wang@duke.edu,
clksong@gmail.com
}

\begin{document}

\maketitle

\begin{abstract}

Word embeddings are traditionally trained on a large corpus in an unsupervised setting, with no specific design for incorporating domain knowledge. This can lead to unsatisfactory performances when training data originate from heterogeneous domains.
%
In this paper, we propose two novel mechanisms for domain-aware word embedding training, namely domain indicator and domain attention, which integrate domain-specific knowledge into the widely used SG and CBOW models, respectively.
The two methods are based on a joint learning paradigm and ensure that words in a target domain are intensively focused when trained on a source domain corpus.
Qualitative and quantitative evaluation confirm the validity and effectiveness of our models.
%
Compared to baseline methods, our method is particularly effective in near-cold-start scenarios.\footnote{Codes are available at \url{https://github.com/guoyinwang/EMBDA}}

\end{abstract}

\section{Introduction}
\label{sec:intro}

Word embeddings have shown their effectiveness in many natural language processing (NLP) tasks \cite{turney2010-JAIR,collobert2011-JMLR,weston2015-ArXiv}. 
To learn embeddings, unsupervised learning is conducted over a large generic corpus, e.g., news text, so as to ensure coverage on various language phenomena and unbiasedness in semantics
\cite{mikolov2013-NIPS,song2018-NAACL,song2017-CoNLL}. 
%
%
Consequently, the resulted embeddings are versatile and can be applied to different tasks, however, losing their characteristics in representing domain-specific information.

Practically, when conducting tasks in a certain domain,
one often has to use the generic embeddings and rely on fine-tuning in later supervised learning \cite{collobert2011-JMLR}.
In this situation, starting with generic embeddings can lead to a tough training process where there exists word distribution mismatch across domains.
\cite{wendlandt2018-NAACL} also confirmed that domain variance greatly affects the quality of word embeddings where their stability within domain is greater than that across domains.
%
%
To obtain domain-specific embeddings,
previous work mainly focuses on capturing domain-specific characteristics 
\cite{bollegala2015-ACL,yang2017-EMNLP}
. 
%
%
%
%

%

Although existing approaches show their effectiveness in learning embeddings by considering domain variances, they are still restricted by relying on massive resources from a target domain.
On the other hand, obtaining sufficient target domain data can be challenging, 
which brings a cold-start issue for
embedding training.
%
For example,
there may be not sufficient training samples that represent different features (e.g., words and contexts etc.) in such a target domain,
%
%
thus affecting regularization and retrofitting approaches,
which rely on the availability of domain-specific data or alternative high quality semantic resources.

To address the aforementioned problem,
we propose an approach for minimizing the impact of target domain data availability. In particular, we aim to learn domain-aware word embeddings by directly exploiting target-domain words and contexts in a sparse corpus.
Two different algorithms are proposed to this end.
The first leverages a domain indicator vector \cite{liu-zhang-liu:2018,nivre:2018} to present domain information in the learning process.
The second uses a domain attention mechanism to highlight those training samples in the source domain that can also be present in the target domain. Distinguishing from  existing methods based on word frequency, both of our methods utilize pairwise word concurrence information, which potentially avoid unitary invariance problem (rotate two words' embeddings without changing distance in between) in shifting word embedding \cite{yin2018-NIPS}.   
%

Experimental results show that our approaches can effectively
learn embeddings with target-specific knowledge,  outperforming embeddings learned solely on a source or target domain, as well as different aggregation methods.
In these tasks, we also demonstrate that our models achieve better results compared to the state-of-the-art methods in previous studies.

\section{Related Work}

Word embeddings have been extensively researched in recent years \cite{collobert2011-JMLR,mikolov2013-NIPS,pennington2014-EMNLP}.
To align embeddings with specific task or application,
there are studies focusing on enhancing embeddings, such as using additional data sources to adapt initial embeddings \cite{maas2011-ACL,yang2015-NAACL},
incorporating external information and refining objective functions
\cite{wieting2015-TACL,nguyen2016-ACL}, and
retrofitting approaches that leverage word relations defined in semantic lexicons or other resources to help adjusting and refining embeddings \cite{faruqui2015-NAACL-HLT,kiela2015-EMNLP}.
Although embedding quality can be improved through external guidance, there is a high barrier from the unavailability of annotated data, especially for domain specific scenarios.



When domain variance is considered,
\citeauthor{bollegala2015-ACL} \shortcite{bollegala2015-ACL} proposed a joint model to learn unsupervised cross-domain word embeddings, assuming that pivot words  frequently shared in both source and target domain have similar behaviors.
However, the assumption does not hold if the source and target domains are much different.
%
%
\citeauthor{yang2017-EMNLP} \shortcite{yang2017-EMNLP} proposed a retrofitting method with frequency driven regularization of shared words across domains.
\citeauthor{xu2018-IJCAI} \shortcite{xu2018-IJCAI} further replace simple word frequency with co-occurrence counts of frequent words in the context, which is similar to our motivation.
However, their model 
directly learns word embeddings on each target domain (task) and hence is unable to deal with domains with scarce data, which is also proved to be a common limitation
for \citeauthor{bollegala2015-ACL} \shortcite{bollegala2015-ACL} and \citeauthor{yang2017-EMNLP} \shortcite{yang2017-EMNLP}.
In comparison, our approach offers an alternative way for learning reliable domain adapted word embeddings when in-domain data is limited.
%


\section{The Approach}



To learn word embeddings with domain awareness,
it requires to capture domain characteristics effectively and smartly,
especially when target domain data is limited.
In general, such domain characteristics are carried by word behaviors (such as distribution, usage, etc.) in different domains.
When targeting on a particular domain, word embedding approaches are expected to enhance the behaviors of the words from that domain so that domain characteristics are learned accordingly.
Conventionally in doing so, words appear in a target domain are emphasized when training them from a source domain, which is confirmed by many previous studies \cite{bollegala2015-ACL,yang2017-EMNLP}.
In this paper, we follow this paradigm so that the learning can be done in an unsupervised manner without other prerequisites.

We propose two models to learn domain aware embeddings based on
%
the widely used skip-gram (SG) and continuous bag-of-words (CBOW) models, which are two algorithms to train word embeddings in \texttt{word2vec} \cite{mikolov2013-NIPS}. 
Our proposed models jointly learn word relations among its context from source and target domain.
To deliver domain-specific knowledge, we use word-word pairs extracted from the target domain to depict word relations in it, where such relations are key component utilized in our models.
Particularly, we have different adaptation strategies for SG and CBOW, regarding to their nature in cooperating with domain-specific knowledge in different ways.
Details are illustrated as follows.

\subsection{Learning with Domain Indicator}

\begin{figure}
\includegraphics[scale=0.65, trim=-20 10 0 10]{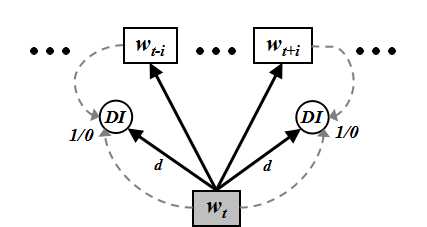}
\caption{Illustration of the SG-DI model. ``\textit{DI}'' refers to the domain indicator, with solid arrows presenting prediction, dashed arrows indicating word-word pair inspection.
}
\label{fig:sgdi}
\vskip -1.5em
\end{figure}

Inspired by \citeauthor{blitzer2006-EMNLP} \shortcite{blitzer2006-EMNLP} and \citeauthor{daume2007-ACL} \shortcite{daume2007-ACL}, words shared in two domains can be leveraged to transfer knowledge across domains.
With word-word pairs extracted from a target domain,
a domain indicator is introduced in this proposed model, namely, SG-DI, for each word in the source domain context to learn whether it is observed with a given word in the target domain.
The proposed model is adapted on the SG model, which is performed 
%
based on the assumption that
a word can be modeled according to
the relations among its neighboring words.
Basically, the SG model predicts the context with a given word,
formulated as maximizing the likelihood
\begin{equation}
\label{equ:sg}
\mathcal{L}_{SG} = \frac{1}{|T|}\sum_{t=1}^{|T|} \sum_{\substack{0< |i|\leq c}} \log f(w_{t+i} \mid w_t), ~~\forall~w_t \in V
\end{equation}
over a corpus $T$,
where $w_t$ and $w_{t+i}$ refer to the given word and context word respectively. $|T|$ is corpus size; $c$ defines the window size.
The $f$ function in this model is conventionally
the probability $p(w_{t+i}|w_t)$ to predict context word, estimated by
\begin{equation}
\label{equ:sg:p}
p(w_{t+i}|w_t) = \frac{exp({\upsilon_{w_{t+i}}^{\prime}}^\top \upsilon_{w_t})}{\sum_{{w_{t+i}}\in V} {exp({\upsilon_{w_{t+i}}^{\prime}}^\top \upsilon_{w_t} )} }
\end{equation}
with $V$ referring to the vocabulary.
Note that for a large vocabulary, SG, as well as CBOW, uses hierarchical softmax or negative sampling \cite{mikolov2013-NIPS} to address the computational complexity requiring $|V|\times dim$ matrix multiplication, where $dim$ is the dimension of word embeddings.

Based on the aforementioned fundamentals,
%
in learning with the domain indicator, a softmax function
\begin{equation}
\label{equ:w2domain}
d(w_{t+i}, w_t) = \frac{exp({\delta_{w_{t+i}}}^\top \upsilon_{w_t})}{\sum_{{w_{t+i}}\in V} {exp({\delta_{w_{t+i}}}^\top \upsilon_{w_t} )} }
\end{equation}
is designed to measure whether a word $w_{t}$ is associated with another word $w_{t+i}$ in the target domain,
by introducing a new vector $\delta$ for each $w_{t+i}$ to present its appearance in the target domain with $w_t$.
The function $d$ shares an updating paradigm similar to negative sampling:
\vskip -1.2em
\begin{align*}
\begin{aligned}
\label{equ:w2lr:update}
\upsilon_{w_t}^{(new)} &= \upsilon_{w_t}^{(old)} - \gamma (\sigma(\upsilon_{w_{t}}^\top \delta_{w_{t+i}}) - \mathcal{D}) \delta_{w_{t+i}}\\
\delta_{w_{t+i}}^{(new)} &= \delta_{w_{t+i}}^{(old)} - \gamma (\sigma(\upsilon_{w_{t}}^\top \delta_{w_{t+i}}) - \mathcal{D}) \upsilon_{w_{t}}
\end{aligned}
\end{align*}
\noindent where $\sigma$ denotes the sigmoid function and $\gamma$ the discounting learning rate.
Particularly, $\mathcal{D}$ is the target label specifying the relative direction of $w_{t+i}$ given $w_t$, defined as
\begin{equation}
\mathcal{D}=\left\{
\begin{array}{rcl}
1       &      & \\
0       &      & \\
\end{array} \right.
\label{eqn:g}
\end{equation}
where $1$ represents that $w_t$ and $w_{t+i}$ co-occurred in the target domain and $0$ for otherwise.
%
The final model is defined as Equation \ref{equ:sg} with $f(w_{t+i}, w_t) = p(w_{t+i}\,|\,w_t) + g(w_{t+i}, w_t)$.

Formally, this model is equal to feature augmentation where the domain indicator plays as the augmented feature specifying the domain variance.
%
Mathematically the validity of doing so was explained by \citeauthor{daume2007-ACL} \shortcite{daume2007-ACL} that each feature has multiple versions for different domains, where the nature of target domain is emphasized when training with such augmented features.
Note that in the extreme case where there is no target domain word pairs utilized in this model, SG-DI becomes the ordinary SG model with $\mathcal{D}=0$ constantly.
Thus in this sense SG-DI can be seen a generalized SG model with considering domain variance.

\subsection{Learning with Domain Attention}

In addition to the ``hard decision'' mode carried out by domain indicator, we also propose another model, namely CBOW-DA, that ``softly'' emphasizes the domain-specific context related to a target domain.
The assumption behind this model is that the words from the target domain that appears in the source domain are considered heavy weighted against other words that only appears in the source domain.
In doing so, words in the context receive a non-uniformed weighting scheme with regards to their domain-specific contributions to the target word.
%
Particularly, this weighting scheme can be seen as a domain attention mechanism for different words where the ones from the target domain are heavier weighted than others.

\begin{figure}
\includegraphics[scale=0.65, trim=-25 10 0 10]{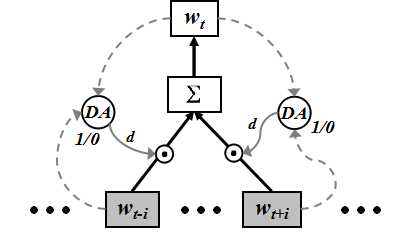}
\caption{Illustration of the CBOW-DA model. ``\textit{DA}'' refers to the domain attention, with black solid arrows presenting prediction,
dashed arrows indicating word-word pair inspection.
$\odot$ is the element-wise production.
}
\label{fig:cbowda}
\vskip -1.5em
\end{figure}
CBOW-DA is constructed based on the CBOW structure,
to formally illustrate our CBOW-DA, we start from the basics of CBOW, 
%
which is different from SG, focuses on maximizing the likelihood that a word is predicted from its context.
%
%
Particularly,
for a given corpus $T$ with vocabulary $V$,
CBOW can be formulated as maximizing the likelihood
\begin{equation}
\label{equ:cbow}
\setlength\abovedisplayskip{8pt} 
\mathcal{L}_{CBOW} = \frac{1}{|T|}\sum_{t=1}^{|T|}\log p(w_t \mid w_{t-c}^{t+c}), ~~\forall~w_t \in V
\end{equation}
where $w_{t-c}^{t+c}$ denotes the context of word $w_t$ in a
window with size $c$.
A projection layer takes the sum of the embeddings from the context words, resulting in $h = \sum_{\substack{0< |i|\leq c}}{\upsilon_{w_{t+i}}}$ over all context words from $w_{t-c}$ to $w_{t+c}$.
%
The probability of predicting $w_t$ with $w_{t-c}^{t+c}$ is defined as
\begin{equation}
\label{equ:cbow:p}
p(w_t|w_{t-c}^{t+c}) = \frac{exp({\upsilon_{w_t}^{\prime}}^\top h)}{\sum_{w_t\in V} exp({\upsilon_{w_t}^{\prime}}^\top h)}
\end{equation}
where $\upsilon_{w_t}$ is the embedding for $w_t$,
and $\upsilon$ and $\upsilon^{\prime}$ refer to input and output embeddings, respectively.

To adapt CBOW with domain attention,
as illustrated in Figure \ref{fig:cbowda}, we introduce an $d(.)$ function to compute the attention assigning to each word pair $w_t$, $w_{t+i}$ with considering its domain information.
Particularly, $d$ is formulated as
\begin{equation}
\label{equ:da}
d = \frac{ exp({\upsilon_{w_t}^{\prime}}^\top \upsilon_{w_{t+i}} ) + k_{w_t,w_{t+i}} }{\sum_{{w_{t+i}}\in V} {[ exp({\upsilon_{w_t}^{\prime}}^\top \upsilon_{w_{t+i}} )} + k_{w_t,w_{t+i}} ]}
\end{equation}
where it is contributed by two parts: $exp({\upsilon_{w_{t+i}}}^\top \upsilon_{w_t})$ indicates how much attention a word $w_{t+i}$ should have in predicting $w_t$, which is a measurement for the word associations in the source domain;
$k_{w_t,w_{t+i}}$ is the domain factor similar to the indicator defined in Eq. \ref{eqn:g}, reflecting if $w_t$ and $w_{t+i}$ co-occurred ($k_{w_t,w_{t+i}} = 1$) in the target domain corpus or not ($k_{w_t,w_{t+i}} = 0$).
The formulation intends to assign words appeared in the target domain with higher weights than others, so that these words contribute more to the learning process.
%
%
With the domain attention,
the projection layer in the updated CBOW model is thus rewritten as
\begin{equation*}
h = \sum_{\substack{0< |i|\leq c}}{d(w_t, w_{t+i})~ \upsilon_{w_{t+i}}}
\label{eqn:att_cbow}
\end{equation*}
%

Followed by the common setting in previous work \cite{bollegala2015-ACL,yang2017-EMNLP} that source and target domain should share some data,
CBOW-DA also assumes there exist enough cases that words appear in both domains.
However, similar to SG-DI, in extreme cases where no target domain words are observed, the proposed attention mechanism still works for each word where $k=0, \forall k$, thus their attention is determined entirely by word associations in the source domain.


\section{Experiments}
\label{sec:exp}

Experiments are done to illustrate the effectiveness of our approach.
The corpus in the source domain that we used to train word embeddings is from the latest articles dumped from Wikipedia\footnote{https://dumps.wikimedia.org/enwiki/latest/} (Wiki for short), which contains approximately 2 billion words.
All the embeddings trained with different approaches share the same hyper-parameters, i.e., 200 dimensions, minimum count of words to 5, a windows size of 5 words, using negative sampling as the learning strategy with 10 negative samples.
These setups are used throughout the following experiments.

In addition to the aforementioned common settings, to facilitate our models, word-word tables are built for each target domain to record word co-occurrences.
To match our setting for embedding training, co-occurred words are only considered within a context window of 5 words, i.e., all word pairs are extracted in a similar way as training embeddings.
In doing so, when word-word pairs from the target domain are found during training on the source domain corpus, they are utilized by SG-DI or CBOW-DA accordingly.

We first conduct qualitative analysis on the generated embeddings when using IMDB corpus \cite{maas2011-ACL} as the target domain dataset, where embeddings learned on different target domain knowledge are compared.
%
Then we employ downstream tasks, i.e., text classification, sequence labeling, to assess the quality of different embeddings.
%
The following baseline models are employed in comparing with ours:
%

\begin{figure}[t]
\includegraphics[scale=0.50, trim=20 10 20 30]{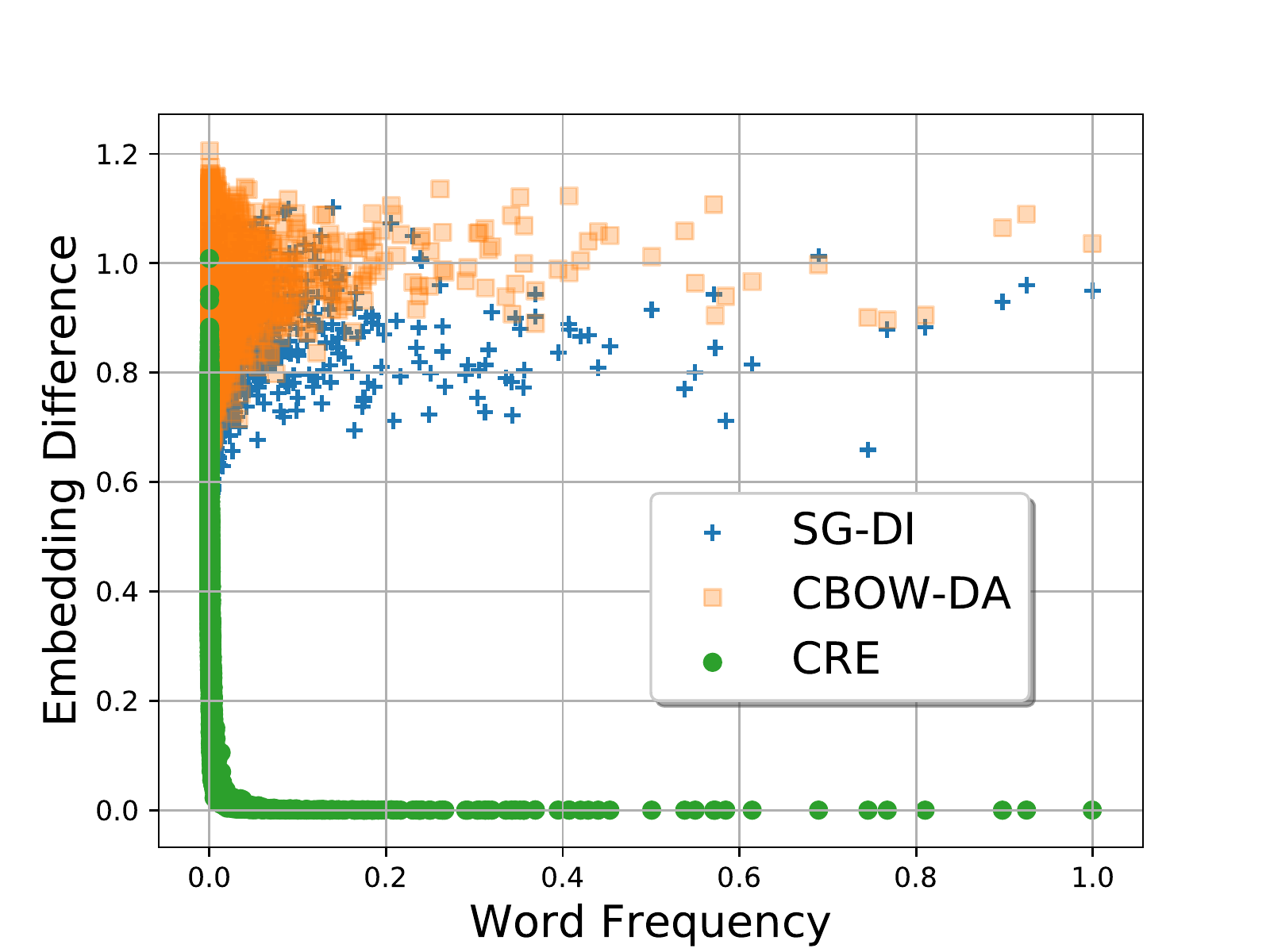}
\caption{Illustration of embedding difference against word frequency.
Embedding difference for each word is measured by cosine distance ($1 - cosine$) between its source and adapted embeddings; all words' frequencies are normalized by S\o rensen-Dice coefficient.
%
}
\label{fig:ed}
\vskip -1.0em
\end{figure}


\begin{itemize}
\item
\textbf{Source embeddings (S)}, CBOW and SG models that trained from the source domain, which is the Wiki corpus in this paper.
\item
\textbf{Target embeddings (T)}, CBOW and SG models that trained from the target domain, which varies in different tasks.
\item
\textbf{Embeddings from all data (+)}, CBOW and SG models are trained from the corpus from both the source and target domain.
\item
\textbf{Concatenation ($\oplus$)}, embeddings that are separately trained from the source and target domain and then concatenated. The concatenation is done between the same typed embeddings, e.g., CBOW S$\oplus$T, SG S$\oplus$T.
\item
\textbf{Averaged (Avg)}, embeddings that are separately trained from the source and target domain and then averaged. Similar to concatenation, average is also done between the same typed embeddings, e.g., CBOW Avg(S,T), SG Avg(S,T).
\item
\textbf{The cross-domain regularized embedding (CRE) method}, which is proposed by \citeauthor{yang2017-EMNLP} \shortcite{yang2017-EMNLP} using a simple regularization algorithm to train cross-domain word embeddings.\footnote{It is reported in \citeauthor{yang2017-EMNLP} \shortcite{yang2017-EMNLP} that the ``DARep'' proposed by \citeauthor{bollegala2015-ACL} \shortcite{bollegala2015-ACL} has a similar setting but significantly worse performance than theirs, we thus do not include DARep in our baselines.}
\item
\textbf{CTCB and CTSG}, the embeddings based on reinforcement learning \cite{song2018-IJCAI} that uses contexts from the target domain to adapt initial word embeddings learned from the source domain. 
\end{itemize}


\begin{figure}[t]
\includegraphics[scale=0.60, trim=20 20 0 0]{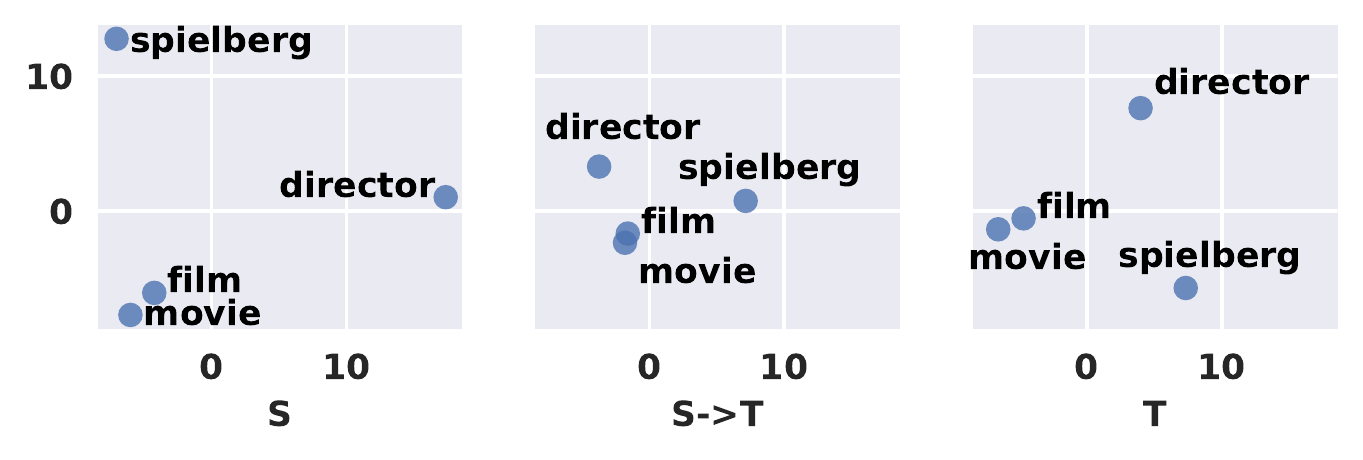}
\caption{Visualization of CBOW embeddings for words \emph{spielberg}, \emph{director}, \emph{film} and \emph{movie}.
From left to right, the embeddings are trained from the source domain (S), cross-domain (S $\rightarrow$ T, i.e., CBOW-DA) and the target domain (T).
}
\label{fig:cbow_v}
\vskip -0.5em
\end{figure}
Once word pairs are prepared, it is straightforward to use these pairs with some retrofitting methods to adapt pretrained embeddings to a target domain.
We add two such methods from \citeauthor{faruqui2015-NAACL-HLT} \shortcite{faruqui2015-NAACL-HLT} and \citeauthor{kiela2015-EMNLP} \shortcite{kiela2015-EMNLP} respectively as additional baselines to illustrate how they are performed in the cross domain setting.

\subsection{Qualitative Analysis}



To demonstrate the benefit of our models from learning word pair information, we first qualitatively compare adaptation effect between our models and others which only utilize frequency information from single words for adaptation.
Herein we choose the state-of-art model, CRE \cite{yang2017-EMNLP}, as the representative of such kind of models for comparison. 

In this analysis,
we use the movie domain and choose IMDB review dataset \cite{maas2011-ACL} as the in-domain corpus, which is a widely used dataset for sentiment classification with 100k positive and negative movie reviews in total.
This dataset contains over 26M word tokens, which can afford training reasonable SG and CBOW embeddings for the target domain.

To effectively quantify embedding shifting, we calculate the cosine distance between embeddings trained on source domain and shifted embeddings from each model. Following the setup in CRE, we use S\o rensen-Dice coefficient \cite{dice1945-Ecology} to represent the normalized word frequency. 

As shown in Figure \ref{fig:ed}, it is observed that embedding trained by CRE has clear separation in shifting: when the frequency of a word is high, its embedding directly follows source domain embedding; when the frequency is low, its embedding remains far away from the source domain and hence barely influenced by the adaptation process.
Moreover,
Figure \ref{fig:ed} also shows that most words in target domain has low normalized frequency.
The CRE method hence does not efficiently utilize source domain information and still suffer from insufficient target domain data.
On the contrary, our models shift word embeddings more evenly over different word frequency.
Particularly noticed from Figure \ref{fig:ed} that,
for words with S\o rensen-Dice coefficient larger than $0.05$, our model consistently shift their embeddings with distances around 0.8 to 1. Such findings indicate that, by utilizing word pair information, our model consistently shift embeddings without significant influence from word frequencies.



To further investigate how word pair knowledge improve the embedding quality for the target domain, we project the embedding vectors to a 2D plane using principal components analysis (PCA) and visualize a group of strong correlated movie-related words. Smaller distance between two nodes on the plots indicates higher relationship. As shown in \ref{fig:cbow_v},  by utilizing word pair information, our model significantly decrease the distance in between and show a more significant cluster. 
Since the sentence representation on many tasks greatly rely on word embedding aggregation\cite{joulin2016-arXiv}, better clustering can provide a better prior knowledge and potentially benefit the downstream tasks.

\begin{table}
\begin{center}
\begin{tabular}{|r|c|c|}
\hline
Embeddings & ~ATIS & ~IMDB~ \\
\hline
\hline
CBOW S & ~~96.19~  & 90.22 \\
\hline
SG S & ~~96.30~  &  90.16\\
\hline
CBOW T & ~~95.74~  & 91.10 \\
\hline
SG T & ~~95.30~  & 90.67 \\
\hline


CBOW S + T & ~~96.19~ & 90.32 \\
\hline
SG S + T & ~~96.30~ &  90.28 \\
\hline

CBOW S $\oplus$ T & ~~96.09~  &  90.55 \\
\hline
SG S $\oplus$ T & ~~95.86~  &  90.87 \\
\hline

CBOW Avg(S,T) & ~~95.86~ &  90.38 \\
\hline
SG Avg(S,T) & ~~95.52~ &  90.45 \\
\hline

CRE & ~~95.86~ &  91.22 \\
\hline

Faruqui et al. \shortcite{faruqui2015-NAACL-HLT} & ~~94.95~ & 89.18 \\
\hline

Kiela et al. \shortcite{kiela2015-EMNLP} & ~~95.41~ & 90.45 \\
\hline
CTCB & ~~97.42~ & 91.09 \\
\hline
CTSG & ~~97.20~ & 90.84 \\
\hline
SG-DI & ~~97.64~ & 91.89 \\
\hline
CBOW-DA & ~~\textbf{97.75}~ & \textbf{92.34} \\
\hline

%

\end{tabular}
\end{center}
\caption{\label{tbl:tc} Text classification results on ATIS and IMDB.
}
\vskip -2.0em
\end{table}


\subsection{Text Classification}
\label{doc_class}
%
The first downstream application to evaluate our embedding models is text classification, where two target domains are considered 
in the experiment.
One task is performed on intent classification in the flight domain, for which we use the ATIS\footnote{ATIS is short for Air Travel Information System.} \cite{Hemphill1990-HLT} corpus for intent classification. The other task is sentiment classification in the movie domain where the
IMDB \cite{maas2011-ACL} corpus used in the previous section is employed for this task. 
The details of datasets are illustrated in Appendix. 
%
Note that, in using ATIS and IMDB, we can better evaluate our proposed models with respect to different sized data in the target domain.

\begin{figure}
\includegraphics[scale=0.26, trim=-40 30 0 50]{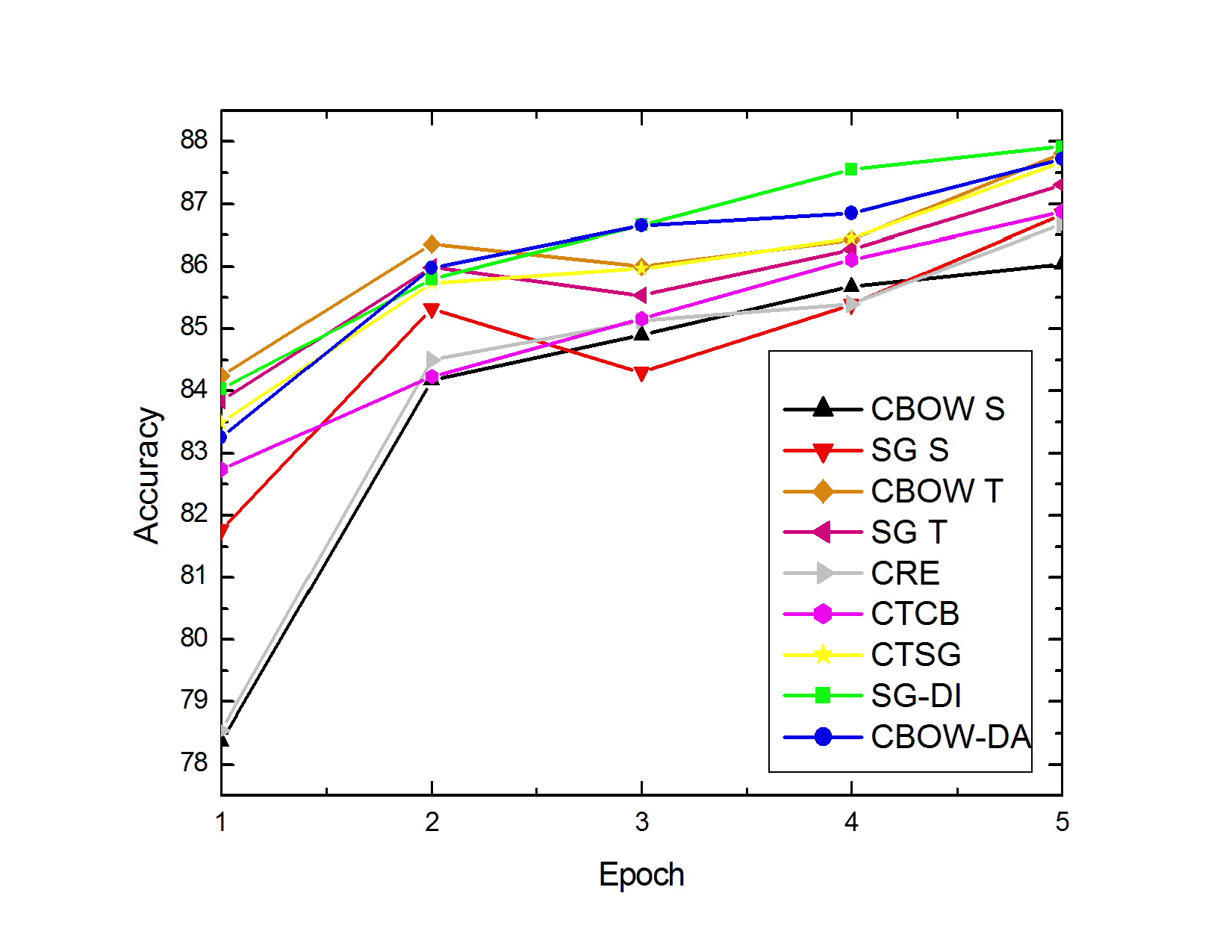}
\caption{Learning curves of the first five epochs on the IMDB task with using different embeddings.
}
\label{fig:curve}
\vskip -1.5em
\end{figure}

For both ATIS and IMDB,
we use a bi-directional LSTM (biLSTM) model to encode each utterance or review for classification.
We have different settings for the hidden states in the biLSTM model for the two tasks.
The ATIS task is performed on a biLSTM with the hidden state size of 256.
For IMDB, because the reviews are longer than usual sentences, we follow the setting of hidden state size 1024 as suggested in \citeauthor{dai2015-NIPS} \shortcite{dai2015-NIPS}.
Embeddings from different models are used as input for the aforementioned classifiers.
In training, all models use cross-entropy loss and Adam optimizer \cite{kingma2014-arxiv} for parameter updating.
Other hyper-parameters (e.g, training epoch and learning rate) are determined by using 15\% of the training data for validation and we apply early stopping strategy when the error rate on the validation set starts to increase.

Results are reported in Table \ref{tbl:tc} for the two tasks with different input embeddings.
%
Overall,
%
in-domain word embeddings tend to outperform out-of-domain ones when in-domain data is relatively large and in a contrary when in-domain data is limited.
%
%
The aggression methods, i.e., S$\oplus$T, Avg(S,T), normally perform in between the embeddings from the source and the target domain;
while the embeddings trained on the concatenated corpus (S+T) tend to perform similar to the ones trained in the source domain because the Wiki corpus dominate the in-domain corpora.
Against all aggression methods and baselines from other studies, our SG-DI and CBOW-DA constantly outperform them with a large margin.
Particularly, no matter whether there is enough in-domain data, SG-DI and CBOW-DA effectively integrate domain knowledge into embeddings and thus result better performance in text classification.
When there are more in-domain data, CBOW-DA demonstrates a larger margin on top of other baselines, as well as SG-DI.
Especially, the performance of CBOW-DA on ATIS achieve a record score 97.75,
which is, to the best of our knowledge, the state-of-the-art result reported on this dataset.
To compare between SG-DI and CBOW-DA, the reason leads to the difference of their performance could be that the attention mechanism works more smooth than the indicator.

\begin{table}[t]
\begin{center}
\begin{tabular}{|r|c|c|c|}
\hline
Embeddings & P & R & F1 \\
\hline
\hline
CBOW S & \textbf{78.69} & 70.75 & 74.51 \\
\hline
SG S & 76.79 & 72.99 & 74.84 \\
\hline
CBOW T & 76.14 & 71.18 & 73.57 \\
\hline
SG T & 76.89 & 71.08 & 73.87 \\
\hline


CBOW S + T & 76.24 & 72.61 & 74.38 \\
\hline
SG S + T & 75.64 & 72.91 & 74.25 \\
\hline

CBOW S $\oplus$ T & 75.00 & 72.97 & 73.97 \\
\hline
SG S $\oplus$ T & 75.47 & 72.97 & 74.20 \\
\hline

CBOW Avg(S,T) & 74.81 & 71.62 & 73.18 \\
\hline
SG Avg(S,T) & 74.91 & 72.13 & 73.49 \\
\hline

CRE & 77.44 & 70.47 & 73.79 \\
\hline

Faruqui et al. \shortcite{faruqui2015-NAACL-HLT} & 72.86 & 69.15 & 70.96 \\
\hline

Kiela et al. \shortcite{kiela2015-EMNLP} & 73.86 & 70.15 & 71.96 \\
\hline

CTCB & 76.52 & 73.03 & 74.73 \\
\hline
CTSG & 75.94 & 72.31 & 74.08 \\
\hline

SG-DI & 77.04 & \textbf{74.18} & \textbf{75.59} \\
\hline
CBOW-DA & 76.76 & 73.30 & 75.51 \\
\hline

%

\end{tabular}
\end{center}
\caption{\label{tbl:genia} NER results on GENIA.
}
\vskip -1.1em
\end{table}

Consider that most complex tasks rely on tuning embeddings in an end-to-end fashion, it is from another point of view to evaluate our embeddings by measuring how they can provide a better staring point for training models for target domain tasks.
In doing so, we take the IMDB task as the case study to show learning curves on test data when training with different embeddings for the first five epochs.
Clearly demonstrated in Figure \ref{fig:curve}, embeddings learned from the target domain have significant better initial performance than that from the source domain.
Although learned from the source domain, SG-DI and CBOW-DA perform similar to the target domain embeddings, while other baseline models are inferior to them.
As a result, a classifier could benefit from less iterations of training when using such embeddings.
%

%
%
%
%
%
%


\subsection{Sequence Labeling}
\begin{table}[t]
\begin{center}
\begin{tabular}{|r|c|c|c|}
\hline
Embeddings & P & R & F1 \\
\hline
\hline
CBOW S & 85.13  & 85.29 & 85.21 \\
\hline
SG S & 84.96 & 84.86 & 84.91 \\
\hline
CBOW T & 79.51 & 79.78 & 79.64 \\
\hline
SG T & 79.55 & 80.18 & 79.86 \\
\hline


CBOW S + T & 85.14 & 85.29 & 85.21 \\
\hline
SG S + T & 85.00 & 84.89 & 84.94 \\
\hline

CBOW S $\oplus$ T & 84.24 & 84.62 & 84.43 \\
\hline
SG S $\oplus$ T & 84.02 & 84.40 & 84.21 \\
\hline

CBOW Avg(S,T) & 84.38 & 84.72 & 84.55 \\
\hline
SG Avg(S,T) & 84.12 & 84.58 & 84.35 \\
\hline

CRE & 81.42 & 81.42 & 81.42 \\
\hline

Faruqui et al. \shortcite{faruqui2015-NAACL-HLT} & 81.12 & 81.52 & 81.32 \\
\hline

Kiela et al. \shortcite{kiela2015-EMNLP} & 82.04 & 82.34 & 82.19 \\
\hline

CTCB & 85.25 & 85.08 & 85.16 \\
\hline
CTSG & 85.23 & 84.85 & 85.04 \\
\hline

SG-DI & \textbf{85.44} & 85.42 & 85.43 \\
\hline
CBOW-DA & 85.41 & \textbf{85.93} & \textbf{85.67} \\
\hline

%

\end{tabular}
\end{center}
\caption{\label{tbl:twitter} POS tagging results on Twitter.
}
\vskip -1.4em
\end{table}

In addition to text classification, we conduct two sequence labeling tasks to evaluate the quality of our embeddings from another type of downstream tasks.
The first one is named entity recognition (NER) on the GENIA v.3.02 dataset \cite{ohta2002-HLT}, which contains 2,000 MEDLINE abstracts, 557,779 tokens, 18,546 sentences and is annotated with 36 fine-grained biological entities.  
The second task is part-of-speech (POS) tagging in the social media domain, which is performed on the ARK dataset \cite{gimpel2011-ACL}, which contains 1,827 tweets with manual POS annotations on English tweets. The preprocessing details of the two datasets are illustrated in Appendix.

Note that, in order to test the effectiveness of our models, we only use the dataset provided in both tasks as the target domain data, which is different from previous studies \cite{bollegala2013-IEEE,yang2017-EMNLP} that external in-domain data are provided.
%
We use a bidirectional LSTM-CRF \cite{huang2015-arXiv,lample2016-NAACL} model as the labeler taking different pretrained embeddings as input.
The LSTM state size is set to 200.
All other settings are the same with the model for text classification.
Precision (P), recall (R) and F1 score are used to evaluate the performance of this task.

The NER results are reported in Table \ref{tbl:genia}.
Overall the embeddings trained in the target domain show worse performance than those trained in the source domain;
all the aggregation baselines perform worse than trained on source domain.
Such findings indicate that with limited data, the quality of embeddings trained on the target domain is not guaranteed.
Interestingly, we find the retrofitting methods obtain the worst results, which show the limitation of such methods in the cold-start scenario with limited data in the target domain.
%
%
%
Compared to other embeddings,
both our models significantly improve the recall and F1 score.
%
To investigate, we notice that the model with our embeddings extract more entities without greatly reducing precision,
while all other embeddings, on the contrary, suffer from precision drop when introducing target domain information. Consider the entities is usually domain-specific terms,
high recall is thus an indicator of how the model is performed with target domain knowledge.
Particularly, our SG-DI and CBOW-DA even outperform the state-of-art layered-BiLSTM-CRF model \cite{ju2018NAACL} 
on GENIA NER with a simple BiLSTM-CRF model.
%

The POS tagging results are reposted in Table \ref{tbl:twitter}.
Owing to the limited target domain data, the overall trend of the results from the source, target and aggregated embeddings is similar to the NER task,
while our SG-DI and CBOW-DA also show improvement 
over all baselines.
Such results illustrate the effectiveness of our models in utilizing and combining the source and target domain information.
Compared to other baseline models, the superiority of our models in both NER and POS tagging may be from the modeling of word-pair relationship, while other models are built from word frequency \cite{yang2017-EMNLP} or context \cite{song2018-IJCAI} from the target domain.
Especially when data target domain is limited, such word-pair information is more robust than others in depicting word-word relations.
%
Moreover, since our SG-DI and CBOW-DA embedding are trained on the source domain corpus, they do not suffer from the data sparsity issue as that in target domain embeddings and those retrofitting methods.

\section{Conclusion}
\label{sec:con}

In this paper, we proposed two domain adaptation methods, namely, domain indicator and domain attention, to enhance the widely used SG and CBOW model with domain awareness.
The two methods leverage word-word relations from the target domain and use them to adjust embedding learning on the source domain.
%
%
Experimental results confirm that, even though trained from the source domain, our embeddings outperform all other baselines in the target domain, especially the embeddings directly from the target domain.
%
More importantly, for scenarios that in-domain data is extremely limited, our proposed models still perform robustly, which illustrate their reliability and adaptability than other methods.

%



\bibliographystyle{named}
\bibliography{bib}
\newpage
\appendix
\section{Experiment Details}
\subsection{Text Classification Setup}
The ATIS corpus contains 5871 utterances with 26 different intents\footnote{It is the conventional setting that in the 26 intents, some of the intents are combinations of multiple single intents.}
such as \emph{flight}, \emph{aircraft}, \emph{airfare}, etc.
The task is thus to assign one of the 26 intents to an given utterance.
In total there are 4978 utterances for training and 893 for testing.
We use all the data in training and test data without labels as the unlabeled data set
for training embeddings from the target domain.

In the total 100k reviews,
the IMDB dataset has 50K labeled ones with equal number of positive and negative reviews,
which is pre-divided into training and test sets, with each set containing 25K reviews.
Identical to the qualitative analysis, we use the entire 100k reviews to train word embeddings from the target domain.

\subsection{Sequence Labeling Setup}
In the NER experiments on GENIA, we follow the standard preprocessing procedure \cite{muis2017-EMNLP}, by splitting the dataset with the first 90\%  sentences for training and the rest for test.
We keep \emph{cell line} and \emph{cell type} entities, and collapse all \emph{DNA}, \emph{RNA}, and \emph{protein} subtypes into \emph{DNA}, and \emph{RNA} , and \emph{protein}, respectively, and remove all other entities.
As a result, there are five entities remaining in the dataset\footnote{The preprocessing follows \citeauthor{muis2017-EMNLP} \shortcite{muis2017-EMNLP}.}. Note that, following \cite{yang2017-EMNLP} setup, we also use the last 10\% of training data as the development set to tune hyper-parameters of our model. 

The ARK dataset has a standard split containing 1,000/327/500 tweets as training/development/test set, respectively.

\end{document}